\documentclass[runningheads]{llncs}

\usepackage{eccv}
\usepackage{eccvabbrv}

\usepackage{graphicx}
\usepackage{booktabs}

\usepackage[accsupp]{axessibility}  

\usepackage{amsfonts}       
\usepackage{nicefrac}       
\usepackage{microtype}      
\usepackage{xcolor}         

\usepackage{caption}
\usepackage{adjustbox}
\usepackage{cite}
\usepackage{amsmath}
\usepackage{amssymb}
\usepackage{epsfig}
\usepackage{algorithm}  
\usepackage{algpseudocode}
\usepackage{pifont}
\usepackage{multirow,multicol, makecell}
\usepackage{bbm}
\usepackage{mathtools}
\usepackage{diagbox}
\usepackage{enumitem}
\usepackage{soul}

\usepackage{tikz}
\usepackage{arydshln}

\newcommand{\mycustomsize}{\fontsize{8}{9.6}\selectfont}

\definecolor{myblue}{RGB}{31,78,121}
\definecolor{mygreen}{RGB}{84,130,53}
\definecolor{myyellow}{RGB}{191,144,0}
\definecolor{myorange}{RGB}{197,90,17}

\newcommand{\gray}[1]{{\textcolor[RGB]{150,150,150}{#1}}}

\usepackage[pagebackref,breaklinks,colorlinks,citecolor=eccvblue]{hyperref}

\usepackage{orcidlink}

\begin{document}

\title{SHERL: Synthesizing High Accuracy and Efficient Memory for Resource-Limited Transfer Learning} 

\titlerunning{SHERL}

\author{
Haiwen Diao\inst{1} \and
Bo Wan\inst{2} \and
Xu Jia\inst{1} \and
Yunzhi Zhuge\inst{1} \and
Ying Zhang\inst{3} \and \\
Huchuan Lu\thanks{Correspondence: lhchuan@dlut.edu.cn. Work was done when Haiwen visited HKUST.}\inst{1} \and
Long Chen\inst{4}
}

\authorrunning{H.~Diao et al.}

\institute{
$^1$Dalian University of Technology \; 
$^2$KU Leuven \; 
$^3$Tencent WeChat \;
$^4$HKUST
}

\maketitle

\begin{abstract}
Parameter-efficient transfer learning (PETL) has emerged as a flourishing research field for adapting large pre-trained models to downstream tasks, greatly reducing trainable parameters while grappling with memory challenges during fine-tuning.
To address it, memory-efficient series (METL) avoid backpropagating gradients through the large backbone. However, they compromise by exclusively relying on frozen intermediate outputs and limiting the exhaustive exploration of prior knowledge from pre-trained models.
Moreover, the dependency and redundancy between cross-layer features are frequently overlooked, thereby submerging more discriminative representations and causing an inherent performance gap (vs. conventional PETL methods).
Hence, we propose an innovative METL strategy called \textbf{SHERL} for resource-limited scenarios to decouple the entire adaptation into two successive and complementary processes.
In the early route, intermediate outputs are consolidated via an anti-redundancy operation, enhancing their compatibility for subsequent interactions; thereby in the late route, utilizing minimal late pre-trained layers could alleviate the peak demand on memory overhead and regulate these fairly flexible features into more adaptive and powerful representations for new domains.
Extensive ablations on vision-and-language and language-only tasks show that SHERL combines the strengths of both parameter and memory-efficient techniques, performing on-par or better across diverse architectures with lower memory during fine-tuning.
Our code is publicly available at: \href{https://github.com/Paranioar/SHERL}{https://github.com/Paranioar/SHERL}.

\end{abstract}

\section{Introduction}
\label{sec:instruction}
Recently, large pre-trained fundamental models have showcased impressive generalization and representation capabilities across various vision~\cite{TransF:ViT,TransF:MAE,TransF:EVA}, language~\cite{TransF:BERT,TransF:GPT-2,TransF:LLaMA}, and multi-modal domains~\cite{VLP:Flamingo,VLP:BLIPv2,VLP:BEiTv3,VLP:EMU,VLM:EVE}.
These models, trained on extensive in-domain benchmarks and online web data, provide a solid starting point for downstream tasks.
Nevertheless, their growing scales make it prohibitively expensive and nearly impossible to fully fine-tune them for specific task adaptation. Hence, parameter-efficient transfer learning (PETL)~\cite{TL:Adapter-BERT,TL:LoRA,TL:Prefix-Tuning} strategies are increasingly favored for domain adaptation to effectively mitigate extreme computational costs and storage usage of model parameters.

\begin{figure*}[!t]
    \centering 
    \includegraphics[width=0.95\linewidth,trim= 0 150 10 0,clip]{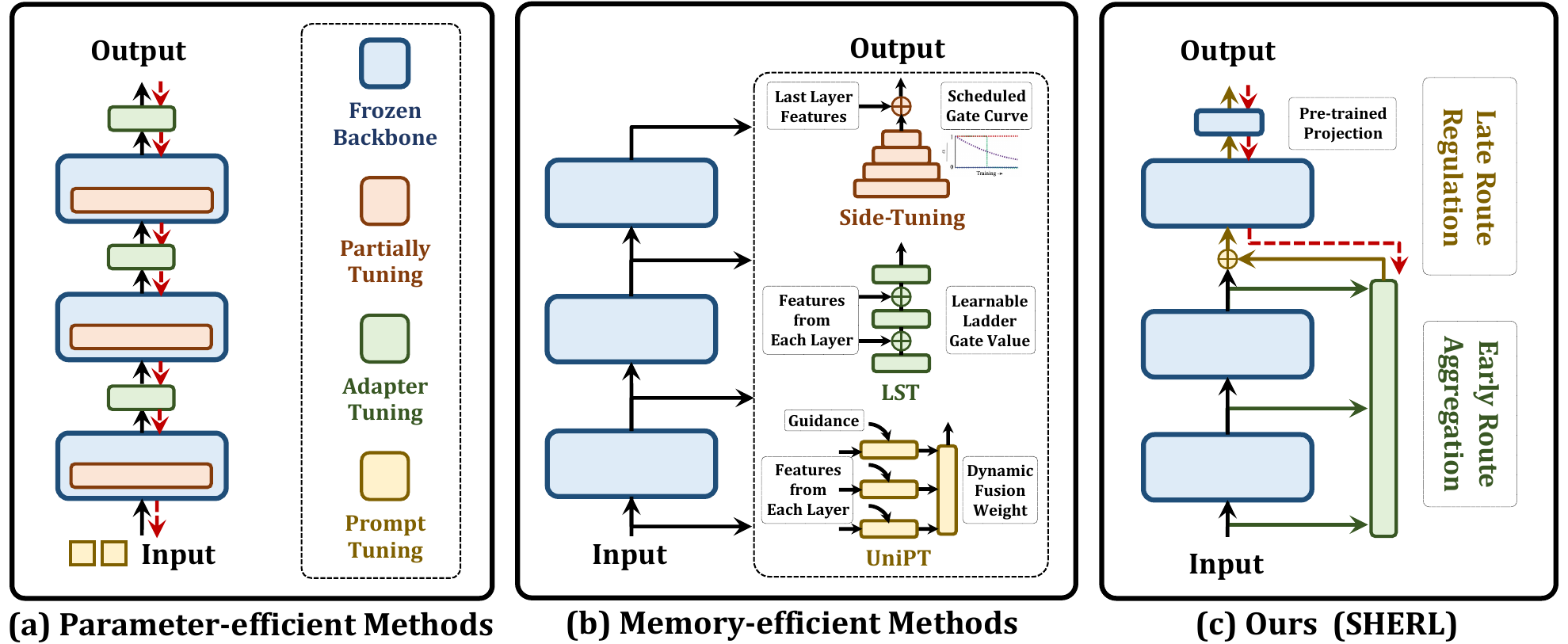}
    \caption{Overview of \textbf{(a)} parameter-efficient
    \textcolor{myorange}{\emph{Partially Tuning}}, 
    \textcolor{mygreen}{\emph{Adapter Tuning}}, 
    \textcolor{myyellow}{\emph{Prompt Tuning}}; 
    and \textbf{(b)} memory-efficient
    \textcolor{myorange}{\emph{Side-Tuning}}~\cite{TL:Side-Tuning}, \textcolor{mygreen}{\emph{Ladder Side Tuning (LST)}}~\cite{TL:LST},
    \textcolor{myyellow}{\emph{Univeral Parallel Tuning (UniPT)}}~\cite{TL:UniPT}. 
    Red dotted line denotes the backward gradients.}
    \label{fig:motivation}
\end{figure*}

Existing PETL methods illustrated in~\cref{fig:motivation}(a) predominantly involve freezing most model parameters and training extra lightweight modules or a few pre-trained layers during fine-tuning. Among them, \textit{Adapter Tuning}~\cite{TL:Adapter-BERT,TL:ViT-Adapter,TL:AdaptFormer} integrates new bottleneck-shaped blocks into the fundamental backbone, while \textit{Prompt Tuning}~\cite{TL:Prefix-Tuning,TL:VPT,TL:CoOp} concatenates a sequence of learnable vectors with the original input, positioned at the beginning or between backbone layers. 
Additionally, \textit{Partially Tuning}~\cite{TL:DiffPruning,TL:FISH-Mask} attempts to update a few task-specific network layers, such as bias items~\cite{TL:BitFit,TL:Tiny-TL} or layer normalization~\cite{TL:Layernorm-Tuning}. 
The PETL strategies significantly decrease training costs and maintain competitive performance as fully fine-tuning. 
Notably, they still incur expensive memory costs due to gradients flowing backward through nearly the entire large backbone.

An alternative research~\cite{TL:Side-Tuning,TL:Y-Tuning,TL:LST,TL:UniPT} in~\cref{fig:motivation}(b) emphasizes more on memory-efficient transfer learning (METL). 
Typically, a parallel side network is constructed alongside the large base network, compromised by exclusively integrating static intermediate features from each layer into the final output. 
However, they overlook two critical problems:
(1) \textit{Redundancy Influence by Cross-layer Aggregation.}
There are diverse dependencies and semantic associations between cross-layer features.
Implicitly, these dependencies categorize corresponding features from different layers into distinct "cohorts" based on their semantic relevance, each cohort containing an indefinite number of cross-layer intermediates. 
Nonetheless, whether the sequential~\cite{TL:LST} or parallel~\cite{TL:UniPT} streamline fails to address feature overlap and redundancy issues during cross-layer aggregation. 
Consequently, early strategies for consolidating these semantic cohorts are indeed biased, favoring the number of layer features within each cohort over their real significance in shaping ultimate representations. 
This is because the cumulative weight of each cohort can be easily swayed by the number of features it contains, potentially diluting small yet meaningful cohorts and jeopardizing the emphasis on each discriminative cohort in the final output.
(2) \textit{Regulation Capability by Pre-trained Layers}.  PETL track~\cite{TL:AdaptFormer,TL:CoCoOp,TL:AdapterFusion} forcibly facilitates feature regulation and migration capability by leveraging the generalized transmission and representation pattern learned by pre-trained parameters.
Conversely, METL track deviates from the original input-output projection of the base network layers and reconstructs the intricate mapping relationship from scratch via lightweight side network layers. 
This choice hinders effective adaptation when translating generic characteristics into domain-specific ones, resulting in representation collapse and suboptimal performance in various downstream applications~\cite{TL:Side-Tuning,TL:LST,TL:UniPT}.

Given the above observations, in this paper, we propose a new tuning paradigm, dubbed \textbf{SHERL}, which deliberately explore a regurgitation-feeding mechanism between early aggregation and late regulation within the pre-trained network (\cref{fig:motivation}(c)). 
Specifically, we detach early tuners from the backbone and commence by quantifying the similarity ratio between cross-layer features, thereby normalizing aggregation weights against the adverse interference of cross-layer redundancy.
Meanwhile, the resulting representations showcase great compatibility with subsequent bare pre-trained layers, which are then transformed into a discriminative space via compulsive regulation.
Our methodology maintains the philosophy of backbone-agnostic unbinding tuner with the “maximalist” pre-trained knowledge under the premise of the “minimalist” training memory overhead.

We evaluate our SHERL on five challenging vision-and-language tasks, (\ie, image-text retrieval~\cite{ITM:SGRAF,ITM:RCAR,ITM:DBL}, video-text retrieval~\cite{Datasets:MSRVTT,Datasets:MSVD}, visual question answering~\cite{Datasets:VQAv2,chen2023counterfactual}, compositional question answering~\cite{Datasets:GQA}, visual grounding~\cite{Datasets:REFCOCO,Datasets:REFCOCOG,chen2021ref}) and a series of pure language-only tasks (\ie, GLUE benchmark~\cite{Datasets:GLUE}).
Extensive ablations show that our SHERL compares favorably with conventional PETL variants with minimal training costs, and meanwhile achieves the optimal balance over METL counterparts between accuracy and efficiency. Besides, SHERL can also seamlessly cooperate with other PETL methods, and display good applicability across various network backbones, encompassing CNN, single or cross-modal Transformer, and T5~\cite{TransF:T5} or MDETR-like~\cite{VLP:MDETR} Encoder-Decoder architectures. 

\section{Related Work}
\label{sec:related-work}

\textbf{Parameter-Efficient Transfer Learning (PETL).}
Large pre-trained models have become essential foundational tools across diverse domains, \eg computer vision (CV)~\cite{TransF:ViT,CL:MoCo,TransF:MAE,TransF:EVA}, natural language processing (NLP)~\cite{TransF:BERT,TransF:RoBERTa,TransF:T5,TransF:GPT-2}, and vision-and-language (VL) applications~\cite{VLP:CLIP,VLP:BLIP,VLP:Imagebind,VLP:meta-transformer}. 
To adapt them for downstream tasks, the standard strategy is to update the entire model parameters during the training process. Nevertheless, as recent foundation models scale up, such an operation becomes extremely expensive and sometimes suffers from sub-optimal issues, especially with small downstream datasets. To address these challenges, parameter-efficient transfer learning (PETL) has attracted increasing research attention, which can be roughly categorized into three directions.
(1) \textit{Partially Tuning.} For example, BitFit~\cite{TL:BitFit,TL:Tiny-TL} exclusively updates bias items of pre-trained layers, while Norm-Tuning~\cite{TL:Layernorm-Tuning} focuses solely on training the layer normalization modules during fine-tuning. Additionally, several methods~\cite{TL:DiffPruning,TL:FISH-Mask} utilize sparse binary masks to control which components of network layers remain static during training. However, owing to the substantial disparity between different architectures and downstream tasks, determining the trainable parts of the pre-trained backbones remains challenging.
(2) \textit{Adapter Tuning.} An alternative technique introduces an extra lightweight and learnable module into the pre-trained backbones, which is widely utilized in uni-modal~\cite{TL:Adapter-BERT,TL:Adapter-NMT,TL:AdapterFusion,TL:CLIP-Adapter,TL:AdaptFormer,TL:ViT-Adapter} and cross-modal domains~\cite{TL:VL-ADAPTER,TL:UniAdapter,TL:VL-PET,TL:VLN-PETL}. They all keep the entire parameters of the pre-trained network frozen during fine-tuning, and the inserted modules can be positioned in parallel or sequentially within the pre-trained network layers~\cite{TL:UnifiedPET,TL:ReLoRA,TL:Convpass}. Among them, some approaches~\cite{TL:SSF,TL:IA3,TL:LoRA,TL:QLoRA} involve scaling and shifting factors, learnable vectors, tiny or quantized multi-layer perception (MLP) modules to refine original feature projection, while other methods~\cite{TL:Compacter,TL:FacT,TL:Hypernetwork,TL:AdaLoRA} further reduce the number of trainable parameters by matrix decomposition and hyper-network prediction for multi-task learning. 
(3) \textit{Prompt Tuning.} The concurrent strategies~\cite{TL:Prefix-Tuning,TL:PEPT,TL:PPT} attempt to insert some sparse manual-tuning~\cite{TL:Frozen,TL:Efficient-Prompt,TL:ScienceQA,TL:MM-CoT,TL:VoP} or dense randomly-initialized~\cite{TL:VPT,TL:CoCoOp,TL:DenseCLIP,TL:MaPLe,TL:PLOT} token features into the original input or each intermediate state of the pre-trained backbone. These pseudo features are learnable and facilitate the adaptation process for specific downstream tasks.
Although they greatly reduce trainable parameters, their memory requirements are still prohibitive, occasionally exceeding those of fully fine-tuning and making them impractical for resource-constrained scenarios.

\textbf{Memory-Efficient Transfer Learning (METL).}
Training memory is predominantly consumed by activations rather than parameters.
In contrast to chasing minimal trainable parameters by the current PETL track, the METL track endeavors to achieve the optimal performance-memory balance under resource-limited scenarios.
(1) \textit{General Memory Optimization.} There are several popular and general techniques~\cite{transF:Reformer,TransF:Rev-ViT} to reduce demand on training memory footprint. For instance, mixed precision training~\cite{Training:MPT} or quantization strategy~\cite{Training:8-bit} introduce mix-precision or low-bit-width formats for weights, activations, and gradients during training. Besides, zero redundancy optimizer~\cite{Training:ZeRO} eliminates memory redundancies in both data- and model-parallel training, enabling efficient scaling of model size with the number of devices, while others avoid storing all intermediate activations and reconstruct discarded ones from backward layers~\cite{CNN:RevNet} or by gradient checkpoint operations~\cite{Training:Sublinear}.
(2) \textit{Decouple Back-propagation from Large Model.} Orthogonally, another research direction emphasizes bypassing gradient computation for the extensive parameters of large models. Typically, some parallel tuners detached from the main backbone facilitate the adaptation process of the pre-trained models with minimal memory overhead.
The simplest manner~\cite{TL:ULMFiT,VLP:CLIP,TransF:T5,CL:SimCLR} is to update an extra projection layer (\eg linear probe) after the last layers of the main backbone. 
Besides, Side-Tuning~\cite{TL:Side-Tuning} introduces an extra lightweight CNN network to augment the static main network for new domains, utilizing an exponential blending schedule.
The output from the last layer is not always optimal. Hence, some studies~\cite{TL:Head2Toe,TL:InCA} opt to select features from one intermediate layer, achieving performance comparable to fine-tuning on out-of-distribution transfer tasks.
However, their minimalist forms inherently restrict the full development of powerful pre-trained representations.
To overcome this, recent works~\cite{TL:VQT,TL:LST,TL:Res-Tuning} introduce query-guided, structure-dependent, or MLP-like side branches from the main branch. They incorporate intermediate features from each layer through lateral connections in a bottom-up sequential manner, while UniPT~\cite{TL:UniPT} propose a modular side module independent from the base backbone and integrates cross-layer hidden states in a top-down parallel fashion.
Unlike UniPT~\cite{TL:UniPT} emphasizing cross-layer semantic alignments guided by the ultimate output, and LST~\cite{TL:LST} constructing input-output transmission via a scaled-down pre-trained backbone, our SHERL takes the first step towards intrinsic redundancy problem during cross-layer aggregation and unexplored regulation ability from pre-trained layers ignored by previous METL works. 
It highlights discriminative features across layers against redundancy interference with minimal memory costs in the early route and meanwhile, leverages pre-trained knowledge from the minimal main layer to maximize transfer benefits in the late route.
Notably, we undergo comprehensive validations on both NLP and VL domains, referencing LST~\cite{TL:LST} and UniPT~\cite{TL:UniPT} respectively, and achieve an optimal accuracy-memory trade-off than other counterparts on two well-established benchmarks.

\section{Methodology}
\label{sec:method}

Commonly, METL track constructs an extra side network to avoid gradients passing through large models, and significantly reduce training memory overhead.
In contrast, PETL track attempts to directly adjust the forward route inside the pre-trained backbone.
They could naturally yield more adaptive representations, thereby resulting in an inherent performance gap than METL variants.

\begin{figure*}[!t]
    \centering 
    \includegraphics[width=0.95\linewidth,trim= 0 410 580 0,clip]{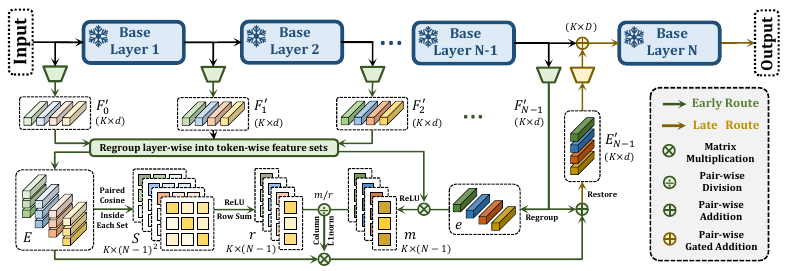}
    \caption{Overview of the framework that mitigates the disparity and redundancy of intermediates across shallow layers, and thereby generates adaptive and compatible inputs for subsequent deep layer for feature regulation when transferred to new domains.}
    \label{fig:framework}
\end{figure*}

Inspired by their respective strengths, we present a new transfer learning paradigm dubbed SHERL in~\cref{fig:framework}, which strikes a sweet balance between model capability and memory efficiency. 
In practice, we categorize all backbone layers into two mutually synergistic groups: $1\to(N - 1)$ shallow layers for early aggregation and $N$-th deep layer for late regulation.
Based on them, we propose an innovative Multi-Tiered Sensing Adapter (MTSA) module encompassing four key components: a down-projection layer set, a non-linear activation layer, a multi-tiered sensing aggregation layer, and a universal up-projection layer. It aims to mitigate cross-layer disparity and redundancy during cross-layer aggregation, and 
bridge compatible connections between shallow and deep backbone layers. 
In~\cref{subsec:early-representation} and~\cref{subsec:late-transmission}, we elaborate on the construction and implementation of our proposed SHERL framework.
In~\cref{subsec:appication_scenarios}, we showcase its versatility and superiority by deploying SHERL across a spectrum of network backbones, including CNNs, Transformers, and Encoder-Decoder architectures.

\subsection{Consolidating Early Feature Representation}
\label{subsec:early-representation}

Firstly, we map $K$ hidden states in $D$ dimension from input embedding and shallow layers into low-dimensional $\boldsymbol{F}=\{\boldsymbol{F}_{n}\in\mathbbm{R}^{K\times d} \, | \, n\in\{0, ..., N-1\}\}$ via $N$ unshared fully-connected (FC) layers and the ReLU activation. Here, $d=D/r$ and $r$ is a reduction factor.
There exists extreme norm variation and value magnitude between the intermediate outputs across layers, stemming from feature heterogeneity between different levels of backbone layers.
To alleviate this issue, we introduce additional learnable layers $\boldsymbol{W}^{\prime}$ for the down-projection features $\boldsymbol{F}$, further enhancing these features using a residual connection. After that, channel-wise factors $\boldsymbol{G}$ are computed from $\boldsymbol{F}$ via extra FC layers $\boldsymbol{W}$ and the Sigmoid function ($\sigma$), each value of which ranges from 0 to 1. They are utilized to modulate the scale magnitude of each feature channel as:
\begin{equation}
\label{eq:weight-scale}
\boldsymbol{G}_{n} = \sigma(\boldsymbol{F}_{n}\boldsymbol{W}_{n}) \, , 
\quad \boldsymbol{F}_{n}^{\prime} = \boldsymbol{G}_{n} \odot (\boldsymbol{F}_{n}\boldsymbol{W}_{n}^{\prime} + \boldsymbol{F}_{n})\, .
\end{equation}
Here $\{\boldsymbol{W},\boldsymbol{W}^{\prime}\} = \{\boldsymbol{W}_{n},\!\boldsymbol{W}_{n}^{\prime}\in\mathbbm{R}^{d\times d} \, | \, n\in\{0, ..., N-1\}\}$ are different trainable matrices for each feature set.
To guarantee the ultimate output for $N$-th deep layer consistent with its original input pattern, we regroup layer-wise feature sets into token-wise ones and adopt the last shallow features as aggregation guidance.
Thereinto, the last feature $\boldsymbol{F}_{N-1}^{\prime}\in\mathbbm{R}^{K\times d}$ is re-organized as late feature set $\boldsymbol{e} =\{\boldsymbol{e}_{k}\in\mathbbm{R}^{1\times d} \, | \, k\in\{1, ..., K\}\}$, while the rest sets $\{\boldsymbol{F}_{i}^{\prime}\in\mathbbm{R}^{K\times d} \, | \, i\in\{0, ..., N-2\}\}$ are reshaped as early feature sets $\boldsymbol{E} =\{\boldsymbol{E}_{k}\in\mathbbm{R}^{(N-1)\times d} \, | \, k\in\{1, ..., K\}\}$. 

After the above permutation, we shift to quantifying the cross-layer dependency inside each feature set.
Concretely, we start by computing the cosine distance within shallow feature $\boldsymbol{E}_{k}$, followed by the ReLU function ($\delta$) to eliminate the irrelevant semantic connections and restrict the matrix scores $\boldsymbol{S}_{k}$ between 0 and 1. 
Note that the diagonal values of $\boldsymbol{S}_{k}$ indicate the cosine similarities between the same hidden features inside $\boldsymbol{E}_{k}$, consistently equating to 1. 
By summing up the rectified values in each row, we obtain the redundancy rate $\boldsymbol{r} =\{\boldsymbol{r}_{k}\in\mathbbm{R}^{(N-1)\times 1} \, | \, k\in\{1, ..., K\}\}$ corresponding to shallow feature sets $\boldsymbol{E}$, where $\boldsymbol{r}_{k}$ indicates overall similarity scores inside $k$-th feature set across layers, \ie the redundancy value of $k$-th hidden state at different shallow layers:
\begin{equation}
\label{eq:redundancy-rate} 
\boldsymbol{S}_{k} = 
\boldsymbol{\tilde{E}}_{k}\boldsymbol{\tilde{E}}_{k}^{\top} \, , \quad 
\boldsymbol{r}_{k} = \delta(\boldsymbol{S}_{k})\mathbf{1} \, .
\end{equation}
Here $\boldsymbol{\tilde{E}}_{k}$ is the normalized $\boldsymbol{E}_{k}$ by row-wise $\ell_2$-norm, and $\mathbf{1}\in\mathbbm{R}^{(N-1)\times1}$ is all-one vector. $\boldsymbol{S}_{k}\in\mathbbm{R}^{(N-1)\times(N-1)}$ is cosine matrix between shallow features in $\boldsymbol{E}_{k}$, and each value of $\boldsymbol{r}_{k}\in\mathbbm{R}^{(N-1)\times1}$ is not less than 1. 
Notably, if the current feature inside $\boldsymbol{E}_{k}$ is not related to the one from any other layer, its redundancy value is equal to 1.
The higher the correlation, the larger the redundancy value.

To satisfy the paradigm and characteristics of the last deep layer, we adopt its original input $\boldsymbol{e}$ as query, and $\boldsymbol{E}$ as key and value. 
To take $k$-th hidden features as an example, we calculate inner-product values rectified by ReLU activation between $\boldsymbol{e}_{k}$ and $\boldsymbol{E}_{k}$, which are then divided by its respective redundancy $\boldsymbol{r}_{k}$ and normalized by $\ell_1$-norm. 
After obtaining the normalized attention weight $\boldsymbol{m}_{k}$, we integrate all the shallow features $\boldsymbol{E}$ with their corresponding  $\boldsymbol{e}$ into the blended features $\boldsymbol{e}^{\prime} =\{\boldsymbol{e}_{k}^{\prime}\in\mathbbm{R}^{1\times d} \, | \, k\in\{1, ..., K\}\}$ as follows:
\begin{equation}
\label{eq:inner_product}
\boldsymbol{m}_{k} = {{\ell_1}\text{-norm}}_{\boldsymbol{E}_{k}} \delta(\boldsymbol{e}_{k}\boldsymbol{E}_{k}^{\top}) \,/\, \boldsymbol{r}_{k}^{\top} \, , 
\quad \boldsymbol{e}_{k}^{\prime} = \boldsymbol{m}_{k}\boldsymbol{E}_{k} + \boldsymbol{e}_{k} \, .
\end{equation}
We adopt linear attention to avoid numerical explosion and unstable gradients of standard softmax attention~\cite{TL:UniPT}. 
The whole complexity of~\cref{eq:redundancy-rate,eq:inner_product} is $K\times(N-1)^{2} + K\times(N-1)$, that is linear with the number of input features.
Implicitly, the final aggregated weights achieve a certain equilibrium for each semantic cohort. For instance, consider two semantic cohorts, $A$ and $A^{\prime}$. The overall weight for $A$ is calculated as $(\sum_{m}a_{m}/r_{m})/(\sum_{m}a_{m}/r_{m} + \sum_{n}a^{\prime}_{n}/r^{\prime}_{n})$, where ($m$, $n$), ($a_{m}$, $a^{\prime}_{n}$), and ($r_{m}$, $r^{\prime}_{n}$) denote the number, attention weight, and redundancy value of cross-layer features in ($A$, $A^{\prime}$). Ideally, $r_{m}$ approaches the cohort size of $A$, and $\sum_{m}a_{m}/r_{m}$ approximates the average weight of features involved in $A$. If we disregard $r_{m}$, the final weight of $A$ becomes susceptible to cohort size ($m$, $n$), \ie adverse interference of feature redundancy, rather than being predominantly dominated by the correlations with the guidance feature.

\subsection{Regulating Late Feature Transmission}
\label{subsec:late-transmission}

After obtaining the merging features $\boldsymbol{e}^{\prime}$, we restore it to the same size of the original input, denoted as $\boldsymbol{E}_{N-1}^{\prime}\in\mathbbm{R}^{K\times d}$ that turns over onto $D$-dimensional representations via a universal up-projection layer. We then import an extra gate parameter $u=\text{tanh}(\frac{\alpha}{T})$, parameterized with a learnable zero-initialized scalar $\alpha$ and temperature $T$ ($= 0.1$) following the default hyperparameters of~\cite{TL:LST}. This gate connection is in charge of controlling the mixing ratio between enhanced features and the original layer input. 
Eventually, these dynamic features automatically adjust themselves to input-output transformation patterns learned by pre-trained layers, as integrated into late layers of the base model.

On top of that, the unbinding property detaches the MTSA module from the base backbone, opening up avenues for diverse 
architecture and deployment possibilities.
One can freely develop more complicated structures for optimal benefits. Meanwhile, the flexible collaborations between early and late routes can seamlessly apply to multiple deep layers tailored to the upper memory limit of current resources.
As mentioned earlier, we aim for “maximalist” pre-trained knowledge under the premise of “minimalist” training memory.
Hence, we connect the MTSA module to the last layer of the base backbone, forming \textit{the standard SHERL} unless specified otherwise. Extensive evidence in~\cref{sec:ablation_studies} shows that such an extremely concise design goes efficiently higher than popular memory-efficient counterparts, and could cooperate with existing PETL methods for better compatibility with early aggregation in a reciprocal manner. 

\begin{figure*}[!t]
    \centering \includegraphics[width=0.99\linewidth,trim= 0 300 330 0,clip]{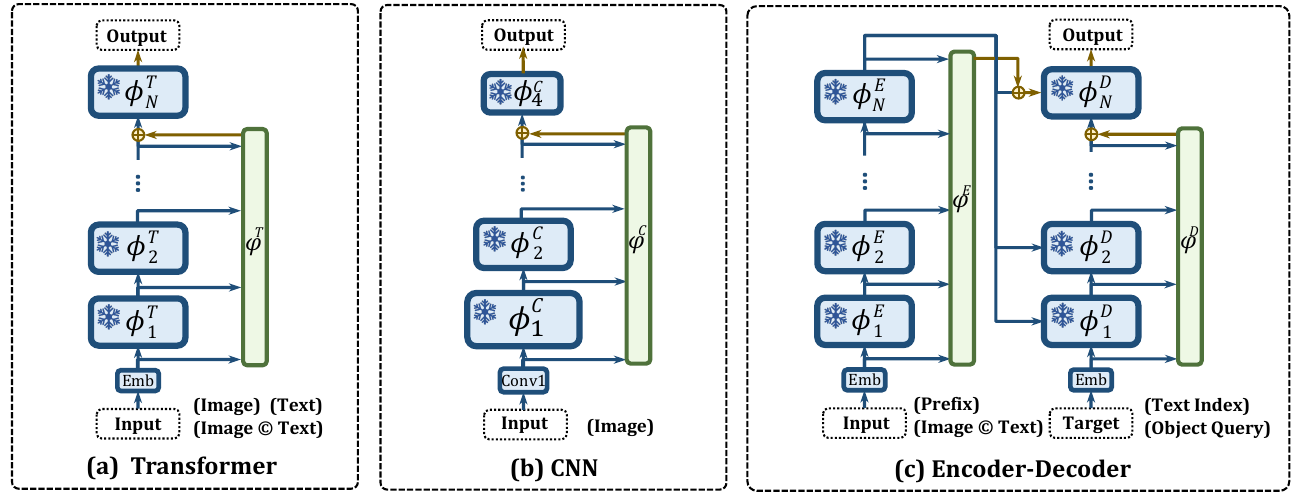}
    \caption{Overview of the Application over (a) single or cross-modality Transformer, (b) CNN, and (c) \emph{T5} or \emph{MDETR}-like Encoder-Decoder architectures. The pre-trained base backbone and our proposed SHERL module are denoted as $\phi$ and $\varphi$, respectively.}
    \label{fig:application}
\end{figure*}

\subsection{Generalized Application Scenarios}
\label{subsec:appication_scenarios}

\textbf{Application on Transformer.}~\cref{fig:application}(a) shows the insertion into various Transformer networks. For single-modality Transformers (\eg BERT~\cite{TransF:BERT} or ViT~\cite{TransF:ViT}), we first extract the word or patch embeddings and shallow intermediate features, which are then consolidated into the final deep layer of the base backbone via the proposed MTSA module.
For cross-modality Transformer (\eg \emph{CLIP-ViL}~\cite{VLP:CLIP-ViL}), image-text pairs are first mapped to the same dimensions and concatenated before being passed to the subsequent cross-modal Transformer $\phi_{1:N}^{T}$ layers. Similarly, input embeddings and hidden states of early base layers $\phi_{1:N-1}^{T}$ are integrated into the last deep backbone layer for domain adaptation.

\noindent\textbf{Application on CNN.}~\cref{fig:application}(b) displays the application with CNN (\eg ResNeXt~\cite{CNN:ResNeXt17}), which utilizes a 2-D convolution kernel as an operator, featuring varied block numbers and stacked structures within each layer. Meanwhile, each feature map from each CNN layer has different spatial and channel sizes (doubled in value). To address this, we first reduce them to the same spatial resolution as the last layer input via several pooling layers. With the inputs of Conv2-5 layers $\phi_{1:4}^{C}$, we follow the same procedure as in Transformer.

\noindent\textbf{Application on Encoder-Decoder.}~\cref{fig:application}(c) exhibits the cooperation with encoder-decoder networks.
They are particular architectures for detection and auto-regressive tasks (\eg \emph{MDETR}~\cite{VLP:MDETR} and \emph{T5-series}~\cite{TransF:T5}). 
We implement the MTSA module in the Encoder-Decoder backbones as its usage in the Transformer. Note that we refrain from feeding the updated features back into the deep layer in the Encoder part to avoid forwarding these layers twice.


\section{Experiments}
\label{sec:experiments}

\subsection{Setup Details}

\textbf{Datasets.}
We validate SHERL on both VL and NLP tasks. For VL tasks, we conduct experiments on \emph{image-text retrieval} (ITR: Flickr30K~\cite{Datasets:Flickr30k}, MSCOCO~\cite{Datasets:MSCOCO}), \emph{video-text retrieval} (VTR: MSVD~\cite{Datasets:MSVD}, MSR-VTT~\cite{Datasets:MSRVTT}), 
\emph{visual and compositional question answering} (VQA: VQAv2~\cite{Datasets:VQAv2}, GQA: GQA~\cite{Datasets:GQA}),
and \emph{visual grounding} (VG: RefCOCO, RefCOCO+~\cite{Datasets:REFCOCO}, RefCOCOg~\cite{Datasets:REFCOCOG}).
For NLP tasks, we adopt GLUE benchmark~\cite{Datasets:GLUE} including \emph{linguistic acceptability} (CoLA~\cite{Datasets:CoLA}), 
\emph{sentiment analysis} (SST2~\cite{Datasets:SST-2}), 
\emph{similarity and paraphrase} (MRPC~\cite{Datasets:MRPC}, QQP~\cite{Datasets:QQP}, STS-B~\cite{Datasets:STS-B}),
and \emph{natural language inference}
(MNLI~\cite{Datasets:MNLI}, QNLI~\cite{Datasets:QNLI}, RTE~\cite{Datasets:RTE}).

\noindent\textbf{Counterparts.}
We compare SHERL against full fine-tuning (\textit{Fully-FT}) and two groups of tuning strategies: \textbf{(1)} Parameter-efficient methods: 
Prompt Tuning (\textit{Prompt}~\cite{TL:Prefix-Tuning}), 
Adapter Tuning (\textit{Adapter}~\cite{TL:Adapter-BERT}, \textit{LoRA}~\cite{TL:LoRA}), 
and Partially Tuning (\textit{BitFit}~\cite{TL:BitFit});
\textbf{(2)} Memory-efficient approaches: \textit{LST}~\cite{TL:LST}
and \textit{UniPT}~\cite{TL:UniPT}.
\textit{Considering their distinct emphasis on parameter and memory requirements, SHERL is primarily compared with memory-efficient LST and UniPT.}

\noindent\textbf{Implementation Details.} We conduct experiments on four NVIDIA TESLA V100 (32GB). To ensure a fair comparison, we adopt the same experimental configurations as employed in LST and UniPT.
For NLP tasks, each method undergoes training with 20 epochs on small datasets (RTE, MRPC, STS-B, CoLA) and 10 epochs for the rest large ones. 
Besides, we set the learning rate and reduction factor to $3\times10^{-3}$ and 8 respectively, following LST~\cite{TL:LST}.
For VL tasks, these values are adjusted to $10\times lr$ and 2 in line with UniPT~\cite{TL:UniPT}, where $lr$ is the learning rate of fully fine-tuning. More details are in Appendix.

\subsection{Main Results}

\textbf{Baselines.} From~\Cref{tab:GLUE_PETL,tab:ITR_METL,tab:VTR_METL,tab:QA_METL,tab:VG_METL}, we demonstrate various transfer platforms and diverse downstream tasks for comprehensive and challenging validations:

\begin{itemize}
    \item \emph{VSE$\infty$}~\cite{ITM:GPO} on ITR task: It utilizes BERT-base~\cite{TransF:BERT} and ResNeXt-101(32×8d)~\cite{CNN:ResNeXt17} pre-trained on Instagram (WSL)~\cite{Datasets:WSL} as text and image backbones. They are pre-trained independently in single-modality domains via a single Transformer or CNN, which then undergoes adaptation for the cross-modality domain.
    \item \emph{CLIP4Clip}~\cite{VLP:CLIP4Clip} on VTR task: It employs pre-trained CLIP~\cite{VLP:CLIP} using Text Transformer~\cite{TransF:GPT-2} and ViT-B/32~\cite{TransF:ViT} models, and adapts these dual Transformer-based encoders from image-text domain to video-text domain.
    \item \emph{CLIP-ViL}~\cite{VLP:CLIP-ViL} on QA task: It adopts CLIP image backbone~\cite{VLP:CLIP} and word embedding layer to encode image and text inputs, followed by a united Transformer for cross-modality interaction. We freeze the pre-trained dual encoders and validate PETL methods on the latter cross-modality Transformer.
    \item \emph{MDETR}~\cite{VLP:MDETR} on VG task: It involves ResNet-101~\cite{CNN:Resnet16} and RoBERTa-base~\cite{TransF:RoBERTa} for image and text encoding respectively, followed by a query-attended encoder-decoder Transformer. We fix the image and text encoding networks and investigate PETL methods on the encoder-decoder structure. 
    \item \emph{T5-series}~\cite{TransF:T5} on GLUE benchmark: they import text encoder and auto-regressive decoder. Following LST~\cite{TL:LST}, we drop 6, 24, 44 layers of side network (3, 12, 22 layers each in encoder and decoder) for \emph{T5-base}, \emph{T5-large}, \emph{T5-3B} to align parameter and memory usage with the baseline.
\end{itemize}

\begin{table*}[t]
    \centering
    \caption{Comparisons with PETL (top) and METL (bottom) methods on GLUE benchmark.
    We utilize \emph{T5-base}, \emph{T5-large}, and larger \emph{T5-3B}. We report accuracy (SST-2, MNLI, QNLI and RTE), Matthew’s Correlation (CoLA), Pearson-Spearman Correlation (STS-B), an average of F1 score and accuracy (MRPC and QQP), respectively.}
    \label{tab:GLUE_PETL}
    \resizebox{\textwidth}{!}{
    \begin{tabular}{lcccccccccccc}
    \toprule
    \multirowcell{2}{Method} 
    &Parameter
    &\multicolumn{2}{c}{Memory (G)} 
    &\multirowcell{2}{CoLA}
    &\multirowcell{2}{SST-2}
    &\multirowcell{2}{MRPC} 
    &\multirowcell{2}{QQP} 
    &\multirowcell{2}{MNLI}
    &\multirowcell{2}{QNLI}
    &\multirowcell{2}{RTE} 
    &\multirowcell{2}{STS-B}
    &\multirowcell{2}{Avg.}\\ 
    &(\%) &Train &Test \\
    \midrule
    \gray{Fully-FT} 
    & \gray{100} & \gray{17.6} & \gray{0.86} 
    & \gray{62.8} & \gray{93.9} & \gray{91.9} & \gray{89.9} & \gray{86.2} & \gray{92.5} & \gray{74.1} & \gray{90.3} & \gray{85.2} \\
    Adapter~\cite{TL:Adapter-BERT} 
    & 1.63 & 13.0 &0.87
    & 64.4 & 94.2 & 88.9 & 88.9 & 86.4 & 93.1 & 75.1 & 91.1 & 85.3 \\
    LoRA~\cite{TL:LoRA}  
    & 1.71 & 12.6 &0.86
    & 63.3 & 94.3 & 90.1 & 89.0 & 86.3 & 93.2 & 75.5 & 90.9 & 85.3 \\
    BitFit~\cite{TL:BitFit} 
    & 0.13 & 10.7 &0.86
    & 61.8 & 94.3 & 91.0 & 88.7 & 85.6 & 93.1 & 67.6 & 90.8 & 84.1 \\
    Prompt~\cite{TL:Prefix-Tuning} 
    & 0.03 & 22.2 &0.87
    & 0 & 90.3 & 74.6 & 88.5 & 82.5 & 92.5 & 59.5 & 90.1 & 72.2 \\
    \midrule
    LST~\cite{TL:LST} 
    &1.74 & 5.5 &0.88
    & 58.1 & 94.1 & 90.4 & 88.8 & 85.6 & 93.3 & 71.9 & 90.7 & 84.1 \\
    \textbf{SHERL} 
    & \textbf{0.85} & \textbf{2.9} &0.87
    & 61.1 & 93.7 & 89.4 & 88.8 & 85.3 & 93.3 & 71.9 & 90.9 & \textbf{84.3}\\ 
    \hdashline
    (\emph{T5-large}) LST~\cite{TL:LST} 
    & 1.23 & 12.2 &2.88
    & 65.3 & 95.7 & 91.6 & 89.7 & 88.6 & 94.1 & 79.9 & 92.4 & 87.1 \\
    (\emph{T5-large}) \textbf{SHERL}
    & \textbf{0.64} & \textbf{7.1} & 2.80
    & 65.6 & 95.8 & 92.9 & 89.6 & 88.6 & 94.2 & 80.8 & 92.1 & \textbf{87.5}
    \\
    \hdashline
    (\emph{T5-3B}) \, \, LST~\cite{TL:LST} 
    & 0.08 & 22.4 & 11.01
    & 66.4 & 96.5 & 92.9 & 89.7 & 90.7 & 95.1 & 80.1 & 93.0 & 88.1\\
    (\emph{T5-3B}) \, \, \textbf{SHERL}
    & \textbf{0.04} & \textbf{21.8} &10.80
    & 67.3 & 96.5 & 92.3 & 89.9 & 90.8 & 95.2 & 80.3 & 92.9 & \textbf{88.2}
    \\
    \bottomrule
    \end{tabular}
    }
\end{table*}

\begin{table*}[ht]
    \begin{minipage}[ht]{\textwidth}
    \centering
    \caption{Comparisons with METL methods on image-text retrieval using \emph{VSE$\infty$} via Recall@1 on sentence (I-T) and image (T-I) retrieval, and Rsum on two directions.}
    \vspace{-8pt}
    \label{tab:ITR_METL}
    \resizebox{\textwidth}{!}
    {\scriptsize
    \begin{tabular}{lcccccccccccc}
    \toprule
    \multirowcell{2}{Method} 
    &Parameter
    &\multicolumn{2}{c}{Memory (G)}
    &\multicolumn{3}{c}{Flickr30K} 
    &\multicolumn{3}{c}{MSCOCO1K} 
    &\multicolumn{3}{c}{MSCOCO5K}
    \\  
    \cmidrule{5-13}
    &(M) &Train &Test 
    &I-T &T-I &Rsum 
    &I-T &T-I &Rsum
    &I-T &T-I &Rsum
    \\
    \midrule
    \gray{Fully-FT} 
    & \gray{201.2} & \gray{22.1 * 8} & \gray{25.15} 
    & \gray{85.6} & \gray{73.3} & \gray{546.6} 
    & \gray{83.1} & \gray{71.7} & \gray{542.7} 
    & \gray{64.2} & \gray{51.2} & \gray{468.9}
    \\
    LST~\cite{TL:LST}  
    &9.7 &24.4 &25.23
    &82.1 &66.5 &529.5
    &78.2 &64.8 &525.8
    &57.8 &43.1 &434.5 
    \\
    UniPT~\cite{TL:UniPT}
    &12.4 &24.4 &25.18
    &84.8 &69.1 &537.4
    &80.6 &67.5 &532.9
    &61.1 &45.9 &445.3
    \\
    \textbf{SHERL}
    &11.3 &24.4 &25.19
    &\textbf{86.1} &\textbf{71.1} &\textbf{542.3}
    &\textbf{81.8} &\textbf{69.2} &\textbf{537.5}
    &\textbf{62.5} &\textbf{47.3} &\textbf{450.8}
    \\
    \bottomrule
    \end{tabular}
    }
    \end{minipage}
    \begin{minipage}[ht]{\textwidth}
    \centering
    \vspace{5pt}
    \caption{Comparisons with METL methods on video-text retrieval using \emph{CLIP4Clip} via Recall@1 on video (T-V) and text (V-T) retrieval, and Rsum on two directions.}
    \vspace{-8pt}
    \label{tab:VTR_METL}
    {\mycustomsize
    \begin{tabular}{lccccccccc}
    \toprule
    \multirowcell{2}{Method} 
    &Parameter
    &\multicolumn{2}{c}{Memory (G)}
    &\multicolumn{3}{c}{MSR-VTT} 
    &\multicolumn{3}{c}{MSVD}
    \\  
    \cmidrule{5-10}
    &(M) &Train &Test
    &T-V &V-T &Rsum 
    &T-V &V-T &Rsum
    \\
    \midrule
    \gray{Fully-FT} 
    & \gray{151.3} & \gray{12.2 * 4} & \gray{1.12} 
    & \gray{42.8} & \gray{42.1} & \gray{389.2} 
    & \gray{45.2} & \gray{57.1} & \gray{425.5}
    \\
    LST~\cite{TL:LST}  
    &11.2 &8.0 * 4 &1.15
    &37.0 &37.8 &356.7
    &35.5 &55.4 &407.2
    \\
    UniPT~\cite{TL:UniPT} 
    &9.6 &3.4 * 4 &1.13
    &38.9 &39.3 &361.3
    &40.9 &59.7 &\textbf{432.1}
    \\
    \textbf{SHERL}
    &9.6 &3.4 * 4 &1.13
    &\textbf{39.2} &\textbf{40.6} &\textbf{363.7}
    &\textbf{40.9} &\textbf{60.2} &429.7
    \\
    \bottomrule
    \end{tabular}
    }
    \end{minipage}
    \\
    \begin{minipage}[ht]{\textwidth}
    \centering
    \vspace{5pt}
    \caption{Comparisons with METL methods on question answering using \emph{CLIP-ViL}.}
    \vspace{-8pt}
    \label{tab:QA_METL}  
    {\mycustomsize
    \begin{tabular}{lccccccc}
    \toprule
    \multirowcell{2}{Method} 
    &Parameter 
    &\multicolumn{2}{c}{Memory (G)}
    &\multicolumn{2}{c}{VQAv2} 
    &\multicolumn{2}{c}{GQA}
    \\  
    \cmidrule{5-8}
    &(M) &Train &Test
    &Test-Dev &Test-Std
    &Test-Dev &Test-Std  
    \\
    \midrule
    \gray{Fully-FT} 
    & \gray{236.8} & \gray{20.5 * 4} & \gray{12.64} 
    & \gray{76.71} & \gray{76.86} & \gray{60.25} & \gray{61.44}
    \\
    LST~\cite{TL:LST}  
    &13.4 &6.4 * 4 &12.76
    &75.29 &75.44 &59.93 &60.75
    \\
    UniPT~\cite{TL:UniPT}
    &10.3 &2.9 * 4 &12.67
    &75.33 &75.53 &60.10 &60.72
    \\
    \textbf{SHERL}
    &13.0 &3.5 * 4 &12.68
    &\textbf{75.53} &\textbf{75.82} 
    &\textbf{60.16} &\textbf{60.82}
    \\
    \bottomrule
    \end{tabular}
    }
    \end{minipage}
    \begin{minipage}[ht]{\textwidth}
    \centering
    \vspace{5pt}
    \caption{Comparisons with METL methods on visual grounding task using \emph{MDETR}.}
    \vspace{-8pt}
    \label{tab:VG_METL}  
    \resizebox{\textwidth}{!}
    {\scriptsize
    \begin{tabular}{lccccccccccc}
    \toprule
    \multirowcell{2}{Method} 
    &Parameter
    &\multicolumn{2}{c}{Memory (G)}
    &\multicolumn{3}{c}{RefCOCO} 
    &\multicolumn{3}{c}{RefCOCO+} 
    &\multicolumn{2}{c}{RefCOCOg}
    \\  
    \cmidrule{5-12}
    &(M) &Train &Test
    &Val &TestA &TestB 
    &Val &TestA &TestB
    &Val &Test 
    \\
    \midrule
    \gray{Fully-FT} 
    & \gray{185.2} & \gray{19.8 * 2} & \gray{3.36} 
    & \gray{86.51} & \gray{89.13} & \gray{81.22} 
    & \gray{79.54} & \gray{84.54} & \gray{70.63} 
    & \gray{80.92} & \gray{80.95}
    \\
    LST~\cite{TL:LST}  
    &0.9 &6.3 * 2 &3.41
    &81.63 &85.19 &76.03
    &71.32 &78.20 &62.06
    &72.53 &73.67
    \\
    UniPT~\cite{TL:UniPT}
    &0.7 &3.4 * 2 &3.38
    &82.71 &86.25 &78.16
    &72.94 &79.18 &64.49
    &77.04 &77.33
    \\
    \textbf{SHERL}
    &0.7 &3.4 * 2 &3.38
    &\textbf{83.02} &\textbf{86.39} &\textbf{78.41}
    &\textbf{73.29} &\textbf{80.11} &\textbf{64.59}
    &\textbf{77.80} &\textbf{77.77}
    \\
    \bottomrule
    \end{tabular}
    }
    \end{minipage}
\end{table*}

\textbf{SHERL outweighs PETL methods under similar training memory.}
\Cref{tab:GLUE_PETL} depicts the performance on the commonly-used GLUE benchmark. Based on \emph{T5-base}, SHERL obtains an 84\% reduction in training memory compared to Fully-FT, while Adapter and LoRA just obtain a 26\% reduction, resulting in SHERL achieving a 3.2x greater memory saving with only half of the parameter usage. Besides, SHERL outperforms BitFit and Prompt with only 27\% and 13\% of their training memory overhead. 
Importantly, the memory-efficient property enables collaboration between SHERL and larger \emph{T5-large}/\emph{T5-3B}, outweighing the baseline with Adapter and LoRA by a large margin under similar training memory conditions.
Notably, compared to Fully-FT, SHERL reduces training time by about 68.8\%, and incurs negligible additional costs of inference time (9.5\%) and memory consumption (0.87GB vs. 0.86GB of the baseline).

\textbf{SHERL outperforms METL competitors in low-memory scenarios.}
We compare SHERL with the best counterparts on NLP tasks in~\Cref{tab:GLUE_PETL}, and observe that it consistently exceeds LST across diverse scales of \emph{T5-series} with only half of trainable parameter usage. More importantly, SHERL saves more than 40\% memory consumption of LST on \emph{T5-base} and \emph{T5-large}.
On five VL tasks, SHERL acquires the closest evaluation results as Fully-FT, and suppresses LST/UniPT by average gains of 4.1/1.1\% in R@1 and 14.1/3.0\% in Rsum on cross-modal retrieval (\Cref{tab:ITR_METL,tab:VTR_METL}), about 0.92/0.65\% improvements on question answering (\Cref{tab:QA_METL}), and total benefits of 20.8/3.3\% on visual grounding (\Cref{tab:VG_METL}).
Note that SHERL exhibits better flexibility and applicability across various challenging architectures under similar training and inference memory consumption as the top competitor UniPT. The solid and consistent improvements across 7 representative backbones and 18 standard datasets confirm the optimal accuracy-memory balance achieved by SHERL than other state-of-the-art counterparts. Besides, SHERL brings an average of roughly 66.8\% training time savings and an extra 13.5\% inference time cost of Fully-FT, showcasing its efficiency and resource-friendly nature under resource-limited scenarios.

\begin{figure*}[t]
    \begin{minipage}[t]{0.48\textwidth}
    \centering
    \includegraphics[width=\linewidth,trim= 0 200 405 5,clip]{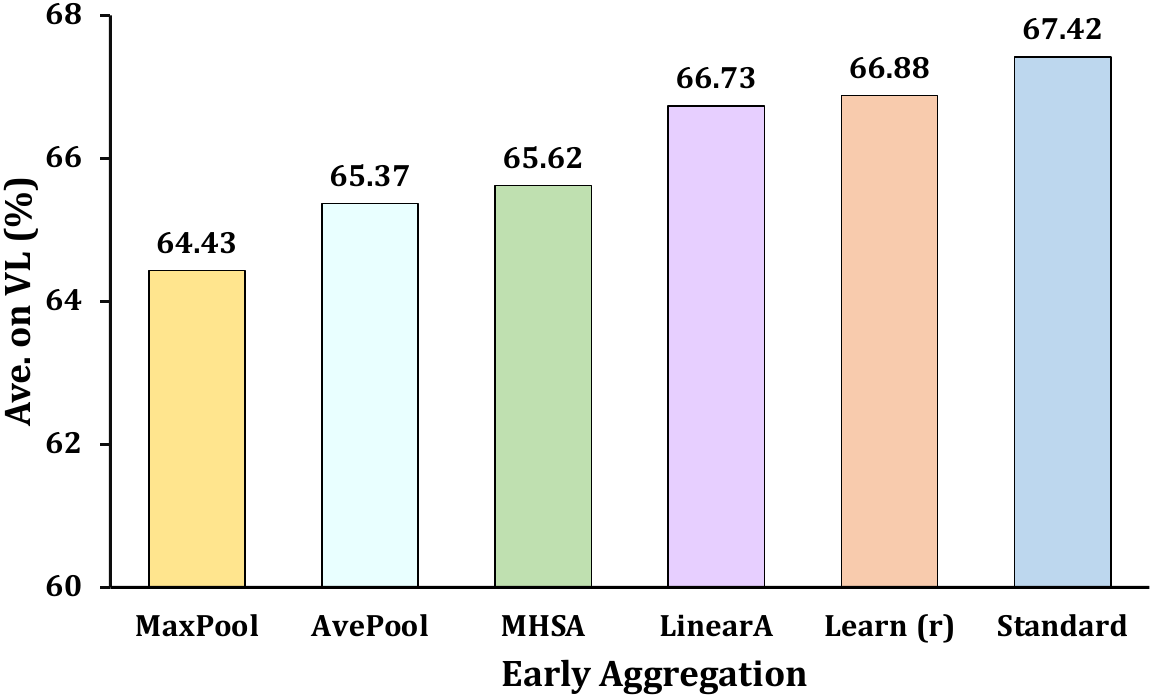}
    \caption{Average accuracy (Ave. \%) on various VL datasets with different aggregation strategies for early intermediate features.}
    \label{fig:redundancy_vl}
    \end{minipage}
    \hfill
    \begin{minipage}[t]{0.48\textwidth}
    \centering
    \includegraphics[width=\linewidth, trim=0 200 405 5,clip]{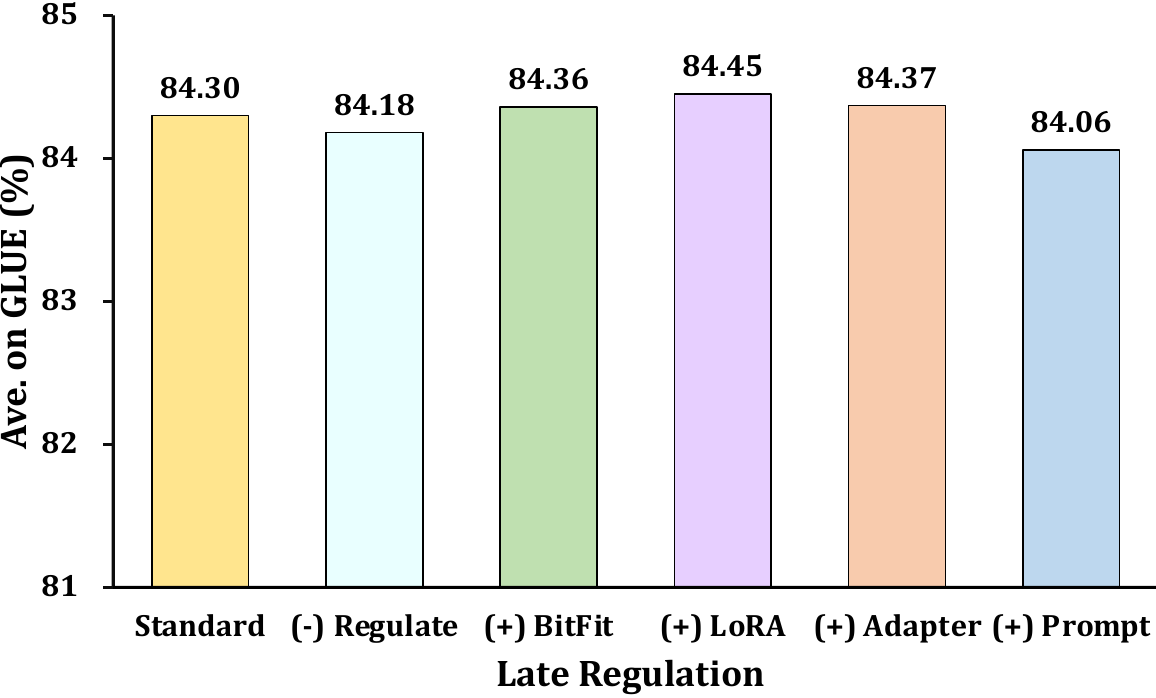}
    \caption{Average accuracy (Ave. \%) on GLUE benchmark by involving popular PETL methods in late feature regulation.}
    \label{fig:regulation_glue}
    \end{minipage}
\end{figure*}

\subsection{Ablation Studies}
\label{sec:ablation_studies}

\textbf{Performance gains through early anti-redundancy and late regulation.}

\underline{\textit{Settings.}} 
In~\cref{fig:redundancy_vl}, we apply SHERL to various pre-trained backbones on diverse VL domains. Here, we replace our early aggregation with various alternative strategies followed by late regulation. Among them, we denote max pooling, average pooling, and multi-head self-attention (head = 4 works best) as \textit{MaxPool}, \textit{AvePool}, and \textit{MHSA}. Besides, we remove the redundancy factor $\boldsymbol{r}_{k}$ in~\cref{eq:inner_product} as plain linear attention (\textit{LinearA}), and replace $\boldsymbol{r}_{k}$ with a learnable parameter initialized by an all-one vector for each early layer as \textit{LinearA (r)}.
In~\cref{fig:regulation_glue}, we validate different late regulation designs of SHERL using \emph{T5-base} on the GLUE benchmark. For all experiments, we keep our early consolidation and incorporate several popular PETL methods into late feature regulation. To hold memory-efficient property, we only insert them into late frozen layers with an extra marginal 0.1-0.2 GB memory (vs 2.9 GB standard). We set the reduction factor of LoRA and Adapter to 16, and the number of prompts to 30. Besides, we treat all base model layers as shallow layers, removing the late regulation and integrating all intermediate features as the ultimate output as \textit{(-) Regulate}.

\underline{\textit{Metrics.}} 
For average accuracy (\textit{Ave.}) on VL domains, we average bi-directional Recall@1 on Flickr30K/MSR-VTT, and accuracy on VQAv2/GQA (Test-Dev) and RefCOCO/+/g (Val) in~\cref{fig:redundancy_vl}. For the GLUE benchmark, we average all the corresponding metrics on eight language-only tasks as \textit{Ave.} in~\cref{fig:regulation_glue}.

\underline{\textit{Results.}} 
From~\cref{fig:redundancy_vl}, we observe that MHSA exceeds simple pooling operations by certain enhancements, while LinearA demonstrates clear advantages. Besides, Learn (r) can adjust the distribution of attention weights to some extent, thereby slightly mitigating redundancy across layers and achieving additional improvements. Compared with them, our MTSA module further obtains notable performance gains, signifying more discriminative representations through such a dynamic anti-redundancy strategy, thus facilitating better domain migration. 
From~\cref{fig:regulation_glue}, exclusively utilizing early fusion without late regulation diminishes the adaptation capability of our SHERL. Moreover, SHERL can effectively incorporate popular PETL strategies and promote the transfer process to downstream tasks, even only with the last layer of the pre-trained backbone. Notably, combining Prompt and SHERL does not yield additional benefits. This is attributed to SHERL already offering sufficient learnable components for learning new domain patterns, causing the network to restrain from updating these prompts with random initialization. Meanwhile, the under-fitting prompts in turn introduce noise interference in the optimization of the proposed MTSA module. 

\begin{figure*}[t]
    \begin{minipage}[t]{0.48\textwidth}
    \centering
    \includegraphics[width=\linewidth,trim= 0 205 350 0,clip]{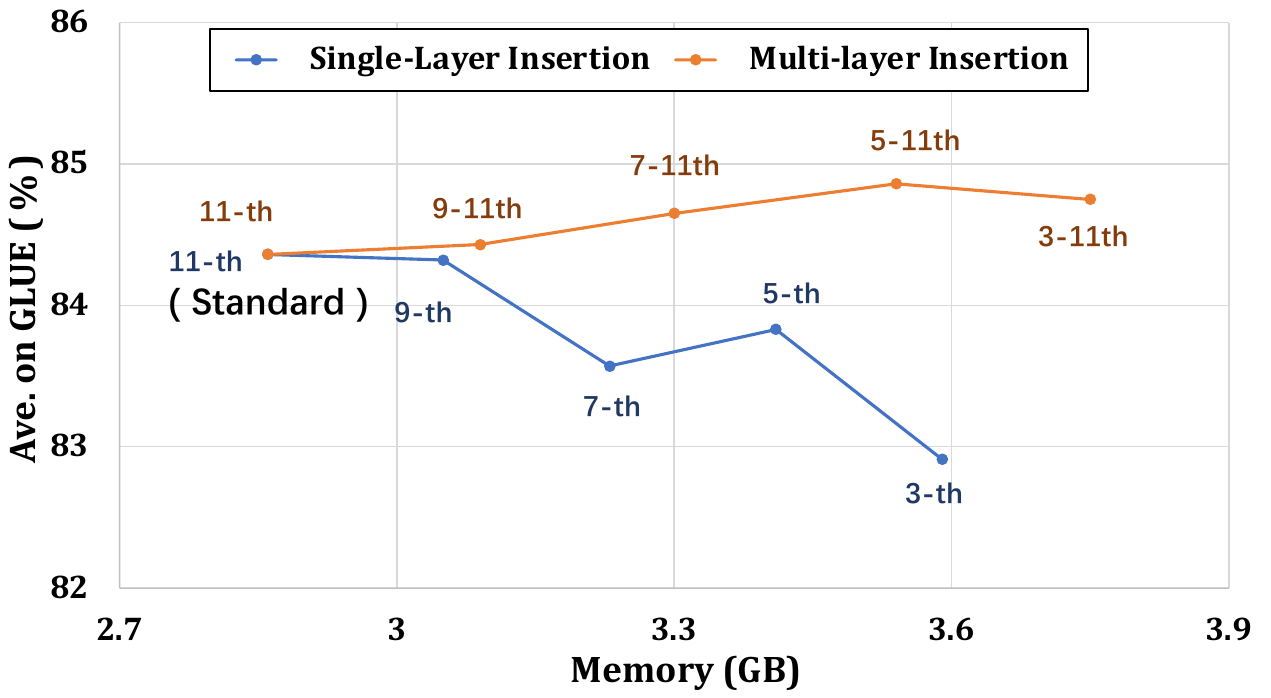}
    \caption{Average results on GLUE benchmark of two insertions into different positions of late decoder layer using \emph{T5-base}.}
    \label{fig:insertion_result}
    \end{minipage} 
    \hfill
    \begin{minipage}[t]{0.48\textwidth}
    \centering
    \includegraphics[width=\linewidth,trim= 0 205 350 0,clip]{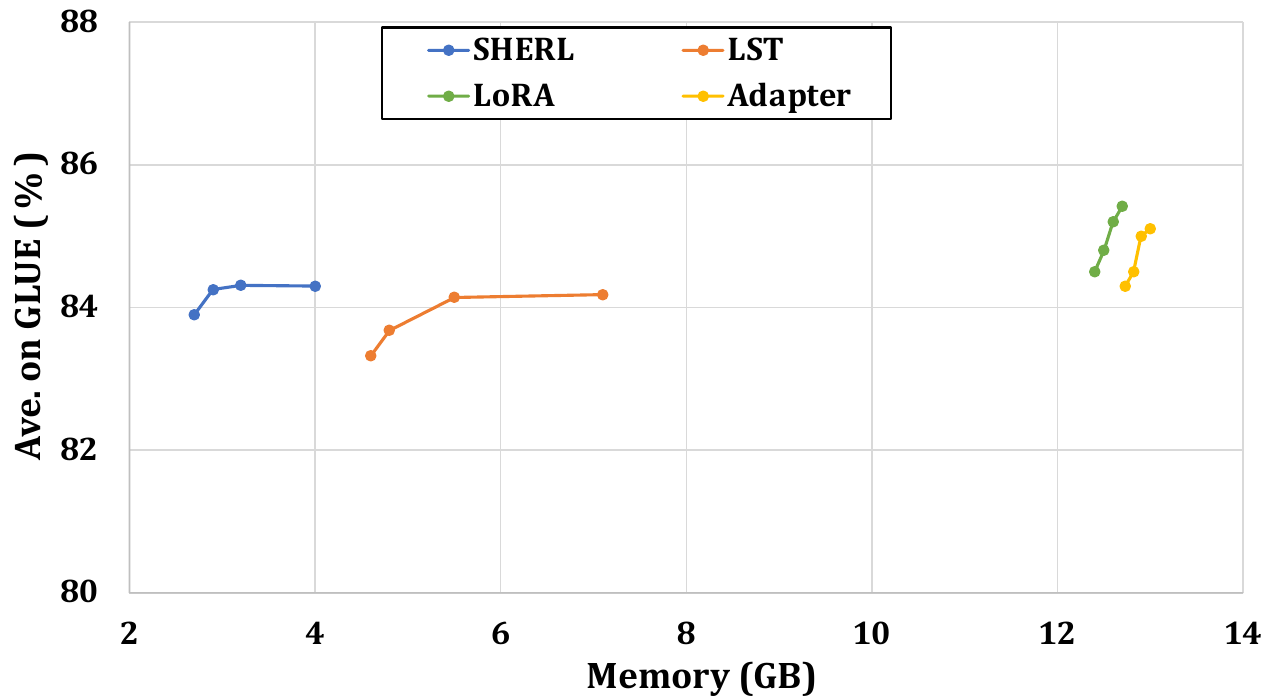}
    \caption{Accuracy-memory trade-off of different strategies via varying reduction factors on GLUE benchmark using \emph{T5-base}.}
    \label{fig:reduction_factor_GLUE_all}
    \end{minipage} 
\end{figure*}

\noindent\textbf{Further benefits with fine-grained insertion patterns and positions.}

\underline{\textit{Settings.}} 
We extend the standard setting (\ie, only the last layer) to single-layer individual or multi-layer collaborative insertion and validate them on the late decoder layers of \emph{T5-base} on GLUE tasks. More details are in Appendix.

\underline{\textit{Results.}} 
In~\cref{fig:insertion_result}, we notice that directly inserting blended features into the earlier decoder layer at once (\ie \textit{Single-layer Insertion}) yields sub-optimal performance. This is primarily due to challenges in unstable gradient flow, making it challenging to optimize the MTSA module. As a solution, we incorporate cumulative feature sets into each late decoder layer (\ie \textit{Multi-layer Insertion}), facilitating multi-level gradient back-propagation for the proposed module. Meanwhile, it flexibly controls the training memory overhead by adjusting the starting position of the late regulation, and dynamically merging previous features to enhance the current input in each late layer. Interestingly, the multi-layer insertion brings additional improvements over the standard setting, showcasing promising potential benefits with further exploration of more complex architectures. 

\noindent\textbf{Optimal trade-off between accuracy and memory over counterparts.}

\underline{\textit{Settings.}} 
To ensure the comparable parameters, we set the hidden size of Adapter as \{6, 12, 24, 48\}, the rank of LoRA as \{4, 8, 16, 32\}, the reduction factor $r$ of LST and SHERL as \{32, 16, 8, 4\} and \{16, 8, 4, 2\}, respectively. Note that for BitFit, the learnable parameters are fixed, and more prompts for prompt tuning suffer from unstable optimization and out-of-memory (OOM) issues.

\underline{\textit{Results.}} 
In~\cref{fig:reduction_factor_GLUE_all}, SHERL outperforms the competitor LST, and achieves the closest results to Adapter and LoRA, meanwhile reducing training memory by a large margin. Besides, SHERL is quite robust across varying $r$, and $r$ = 8 can gain the best trade-off with half the parameters of LST, Adapter, and LoRA. 

\begin{figure*}
    \begin{minipage}[t]{0.48\textwidth}
    \centering
    \includegraphics[width=\linewidth, trim= 2 270 591 2,clip]{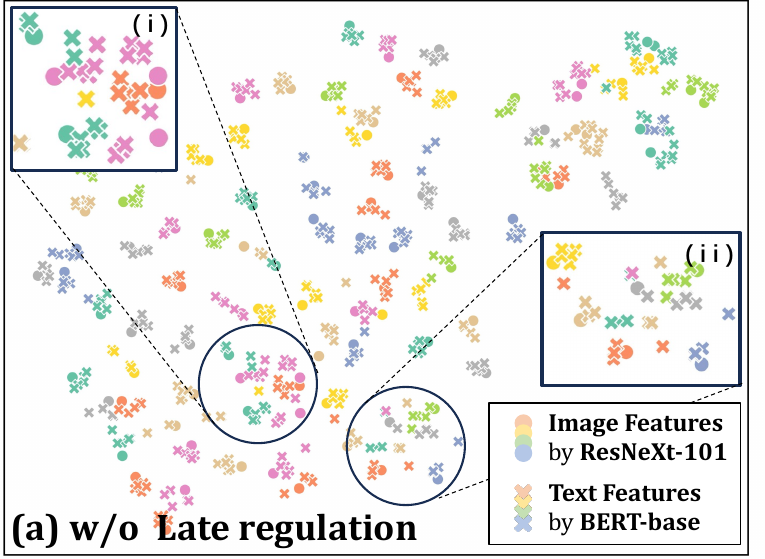}
    \end{minipage}
    \hfill
    \begin{minipage}[t]{0.48\textwidth}
    \centering
    \includegraphics[width=\linewidth,trim= 0 270 593 2,clip]{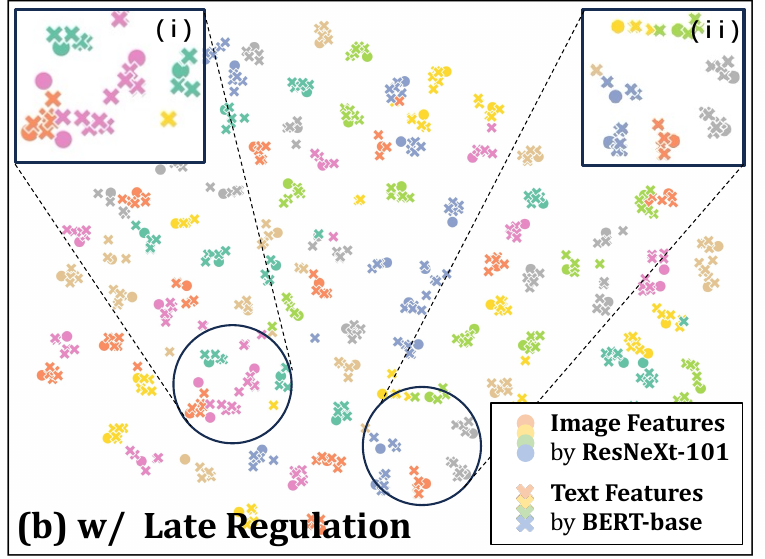}
    \end{minipage}
    \caption{Distributions of the ultimate features through \emph{VSE$\infty$} on Flickr30K by t-SNE.}
    \label{fig:visualization_regulation}
\end{figure*}

\noindent\textbf{Qualitative analysis of regulation ability through late pre-trained route.}

\underline{\textit{Settings.}} 
In~\cref{fig:visualization_regulation}, we randomly select 100 images and 500 related captions on the Flickr30K test set. Based on \emph{VSE$\infty$}, we set (n\_components = 2, learning\_rate = 'auto', init = 'pca') for t-SNE from the sklearn package to show feature distributions by CNN and Transformer with or without late regulation.

\underline{\textit{Results.}} Comparing the feature distributions within areas (i) and (ii), we discover that for both CNN and Transformer networks, late regulation could enhance feature separation, especially for challenging image-text pairs. 
Such effectiveness stems from following the pre-trained input-output projections, and smoothly adapting generic patterns into domain-specific characteristics through forcible constraint and guidance. Consequently, this process generates more discriminative and powerful representations for new domain adaptation.

\section{Conclusion}
In this paper, we propose a brand-new METL platform dubbed SHERL, aiming to acquire outstanding performance and efficient memory in resource-limited applications. 
Specifically, it involves an unbinding Multi-Tiered Sensing Adapter (MTSA) from the pre-trained model, effectively consolidating early intermediate outputs across layers against redundancy problems, and displaying great compatibility with late compulsive regulation in a reciprocal manner.
With the characteristics of METL track, SHERL does not require back-propagation through the large backbone network, and significantly reduces training memory overhead for accommodating much larger and stronger networks, outperforming popular PETL strategies. 
Extensive experiments validate its good flexibility and broad applicability across various network structures including CNN, Transformer, and Encoder-Decoder architectures.
Last but not least, SHERL further exhibits promising potential by incorporating existing PETL methods or more complicated designs, providing a better trade-off between performance gain and memory usage.
We hope that SHERL can aid users with limited computational resources in fine-tuning large models across diverse domains.

\noindent\textbf{Discussion.} We view SHERL as a starting point to bridge and combine the strengths of parameter/memory-efficient approaches for better resource-limited transfer learning. Besides, SHERL holds further potential by undergoing various morphing and combining paradigms with more pre-trained backbone layers as the parameter-efficient version. To maintain the original intention of ``minimalist'' resource costs, further exploration of potential variants and collaboration with more popular and larger foundational models (\eg, LLMs and vision backbones for understanding and generation) are reserved for future research.

\begin{appendix}
\noindent In the Appendix, more details and experiments are organized as follows:
\begin{itemize}[leftmargin=0.4cm, itemindent=0.4cm]
    \item Ethics Statement;
    \item Datasets and Metrics;
    \item Implementation Details;
    \item Multi-layer Insertion Extension;
    \item Hyper-parameter Reduction Factor.
\end{itemize}

\section{Ethics Statement}
We introduce SHERL, an initial exploration into parameter- and memory-efficient strengths, and develop efficient fine-tuning strategies for pre-trained networks in the VL and NLP domains. We note that ethical considerations surround large pre-trained models, including potential bias or discrimination~\cite{ES:ETD_LLM}, and privacy concerns~\cite{ES:IRGB} in the extensive training data. Moreover, computational cost and environmental influence have become increasingly the topics of discussion~\cite{ES:EPC_NLP}.

To our knowledge, previous PETL methods have not delved into the investigation of whether they exacerbate, alleviate, or negligibly affect concerns like bias or information leakage, which needs further exploration. 
When compared with fully fine-tuning, SHERL trains large pre-trained models with negligible tuning parameters and fewer computing resource requirements during the training and inference. This reduction in model deployment costs encompasses both memory and server resources. Lastly, it is noteworthy that our training experiments are conducted in a data center powered entirely by renewable energy sources.

\section{Datasets and Metrics}

\textbf{Image-Text Retrieval (ITR)}: We employ MSCOCO~\cite{Datasets:MSCOCO} and Flickr30K~\cite{Datasets:Flickr30k} for image-text matching task. Note that each image in these datasets corresponds to five text descriptions.
There are 123,287 and 31,000 images on the MSCOCO and Flickr30K datasets, respectively. Following the standard split~\cite{ITM:VSE++}, we partition MSCOCO into 113,287 training images, 5000 validation images, and 5000 test images. For Flickr30K, we divide it into 29,000 training images, 1000 validation images, and 1000 test images.
Besides, the evaluation results on the MSCOCO dataset are presented for both the 5K and 1K test sets. Specifically, for the 1K test set, the reported results are averaged across five distinct 1K folds.

\textbf{Video-Text Retrieval (VTR)}: We import MSR-VTT~\cite{Datasets:MSRVTT} and MSVD~\cite{Datasets:MSVD}.
MSR-VTT contains 10,000 videos, each ranging from 10 to 32 seconds. Each video has 20 related caption annotations. Following the standard split~\cite{VLP:MMT}, we employ the 1k-A protocol, where 9,000 videos with all corresponding captions are for training, and 1,000 video-text pairs are for testing.
Moreover, MSVD contains 1,970 videos with approximately 80,000 captions, where 1,200, 100, and 670 videos are utilized in train, validation, and test splits, respectively. We present test results considering multiple captions per video.

\textbf{Question Answering (VQA\&GQA)}: VQAv2~\cite{Datasets:VQAv2} comprises 83k / 41k / 81k images with 443k / 214k / 453k question-image pairs for training / validation / testing split, while GQA~\cite{Datasets:GQA} consists of 113K images and 22M questions generated from ground truth image scene graphs. Notably, we validate performance on both test-dev and test-standard splits through the standard EvalAI system.

\textbf{Visual Grounding (VG)}: we adopt RefCOCO, RefCOCO+~\cite{Datasets:REFCOCO}, and RefCOCOg~\cite{Datasets:REFCOCOG} for validation. 
RefCOCO has 142,210 expressions for 50,000 bounding boxes in 19,994 MSCOCO images. These are categorized into train, validation, Test A, and Test B, with 120,624, 10,834, 5,657, and 5,095 samples, respectively. Test A focuses on bounding boxes containing people, while Test B involves objects.
RefCOCO+ has 141,564 expressions for 49,856 boxes in 19,992 MSCOCO images, divided into train (120,191), val (10,758), Test A (5,726), and Test B (4,889) splits. Its expressions include more attributes than absolute locations.
RefCOCOg has 104,560 expressions for 54,822 objects in 26,711 images. The expressions are collected in a non-interactive fashion, resulting in an average length of approximately 8.4 words, which is longer.

\textbf{GLUE Benchmark}~\cite{Datasets:GLUE}: We adopt eight standard NLP tasks, consisting of linguistic acceptability (CoLA~\cite{Datasets:CoLA}), sentiment analysis (SST2~\cite{Datasets:SST-2}), similarity and paraphrase (STS-B~\cite{Datasets:STS-B}, MRPC~\cite{Datasets:MRPC}, QQP~\cite{Datasets:QQP}), and natural language inference (RTE~\cite{Datasets:RTE}, QNLI~\cite{Datasets:QNLI}, MNLI~\cite{Datasets:MNLI}).
We report the Accuracy metric for SST-2, MNLI, QNLI, and RTE.
For CoLA and STS-B, we present Matthew's Correlation and Pearson-Spearman Correlation as evaluation metrics, respectively.
Besides, we average the F1 score and Accuracy metrics for MRPC and QQP. Notably, we report the maximum training and testing memory usage on the RTE dataset, and compute the average results with three different seeds.

\begin{table*}[t] \scriptsize
    \centering
    \caption{Hyperparameters of PETL and METL methods on the GLUE benchmark.}
    \label{tab:hyperparameters-glue-tasks}
    \begin{tabular}{lccc}
    \toprule
    Method
    &\quad Learning rate \quad
    &\quad Batch size \quad 
    &\quad Other hyper-parameters \quad \\             
    \midrule
    Fully-FT
    & 3e-4 & 100 & -- \\
    Adapter 
    & 3e-4 & 100 & hidden dimension=48\\
    LoRA  
    & 3e-4 & 100 & rank=32 \\
    BitFit  
    & 3e-4 & 100 & -- \\
    Prompt
    & 3e-1 & 100 & prompt number=100 \\
    \midrule
    LST (\emph{T5})
    & 3e-3 & 100 & \makecell{reduction factor=8}\\
    SHERL (\emph{T5})
    & 3e-3 & 100 & \makecell{reduction factor=8}\\
    \bottomrule
    \end{tabular}
\end{table*}

\begin{table*}[t]
    \centering
    \caption{Hyperparameters of SHERL on ITR, VTR, VQA, GQA, and VG tasks.}
    \label{tab:hyperparameters-vl-tasks}
    \resizebox{\textwidth}{!}{
    \begin{tabular}{lccccccccc}
    \toprule
    Settings
    &MSCOCO &Flickr30K
    &MSR-VTT & MSVD
    &VQA &GQA
    &RefCOCO &RefCOCO+ &RefCOCOg \\
    \midrule
    Learning rate 
    &5e-4 &5e-4 &1e-4 &1e-4 &5e-4 &1e-4 &5e-4 &5e-4 &5e-4\\
    Total epochs 
    &25 &25 &5 &5 &5 &5 &10 &10 &10\\
    Warmup 
    &linear &linear &cosine &cosine
    &linear &linear &linear &linear &linear\\
    Batch size 
    &112 &112 &128 &128 &256 &256 &8 &8 &8\\
    AdamW $\beta_{1}$ 
    &0.9 &0.9 &0.9 &0.9 &0.9 &0.9 &0.9 &0.9 &0.9\\
    AdamW $\beta_{2}$
    &0.999 &0.999 &0.98 &0.98
    &0.999 &0.999 &0.999 &0.999 &0.999\\
    Weight decay 
    &1e-2 &1e-2 &1e-2 &0.0 
    &1e-2 &1e-2 &0.0 &0.0 &0.0\\
    Vision resolution 
    &512$^{2}$ &512$^{2}$ &12*224$^{2}$ &12*224$^{2}$
    &384*640 &384*640 &raw size &raw size &raw size\\
    \bottomrule
    \end{tabular}}
\end{table*}

\section{Implementation Details}
In~\cref{tab:hyperparameters-glue-tasks,tab:hyperparameters-vl-tasks}, we illustrate the intricate configurations of hyper-parameters for both the GLUE benchmark and VL tasks. It's worth noting that on the GLUE benchmark, we adhere to the drop layer strategy outlined in LST~\cite{TL:LST}. This entails discarding the 0th, 4th, and 8th encoder and decoder layers in the T5-base model. When it comes to the T5-large model, it involves dropping the even-indexed encoder and decoder layers. In terms of the T5-3B model, only the 22nd and 23rd encoder layers are retained for feature aggregation and regulation.

\begin{figure*}[h]
    \begin{minipage}[t]{0.96\textwidth}
        \centering
        \includegraphics[width=0.48\linewidth,trim= 0 445 710 2,clip]{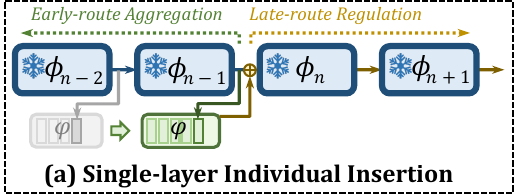} 
        \hfill
        \includegraphics[width=0.48\linewidth,trim= 0 348 710 98,clip]{figures/insertion_style.pdf}
        \caption{Illustration of feeding early-route merging features into the late-route regulation via two types of insertion, including single-layer or multi-layer insertion.
        }
        \label{fig:insertion_style}
    \end{minipage}
\end{figure*}

\section{Multi-layer Insertion Extension}

In~\cref{fig:insertion_style}, we extend the standard setting (\ie, only the last layer) to explore more insertion fashions. Here, we validate them with the late decoder layer of T5-base on the GLUE benchmark. Directly inserting blended features into the earlier decoder layer at once (\ie \textit{Single-layer Insertion}) yields sub-optimal performance, due to the challenges in unstable gradient flow and optimization of the MTSA module. Hence, we incorporate cumulative feature sets into each late layer (\ie \textit{Multi-layer Insertion}), facilitating multi-level gradient back-propagation. Meanwhile, it flexibly controls the training memory overhead by adjusting the starting position of the late route, dynamically leveraging all previous features to enhance late regulation. Notably, multi-layer insertion results in further enhancements beyond the standard setting outlined in the original main body, underscoring its promising potential for continued exploration and development.

\begin{figure*}[t]
    \begin{minipage}[t]{0.48\textwidth}
        \centering
        \includegraphics[width=\linewidth, height=0.65\linewidth,trim= 0 185 345 0,clip]{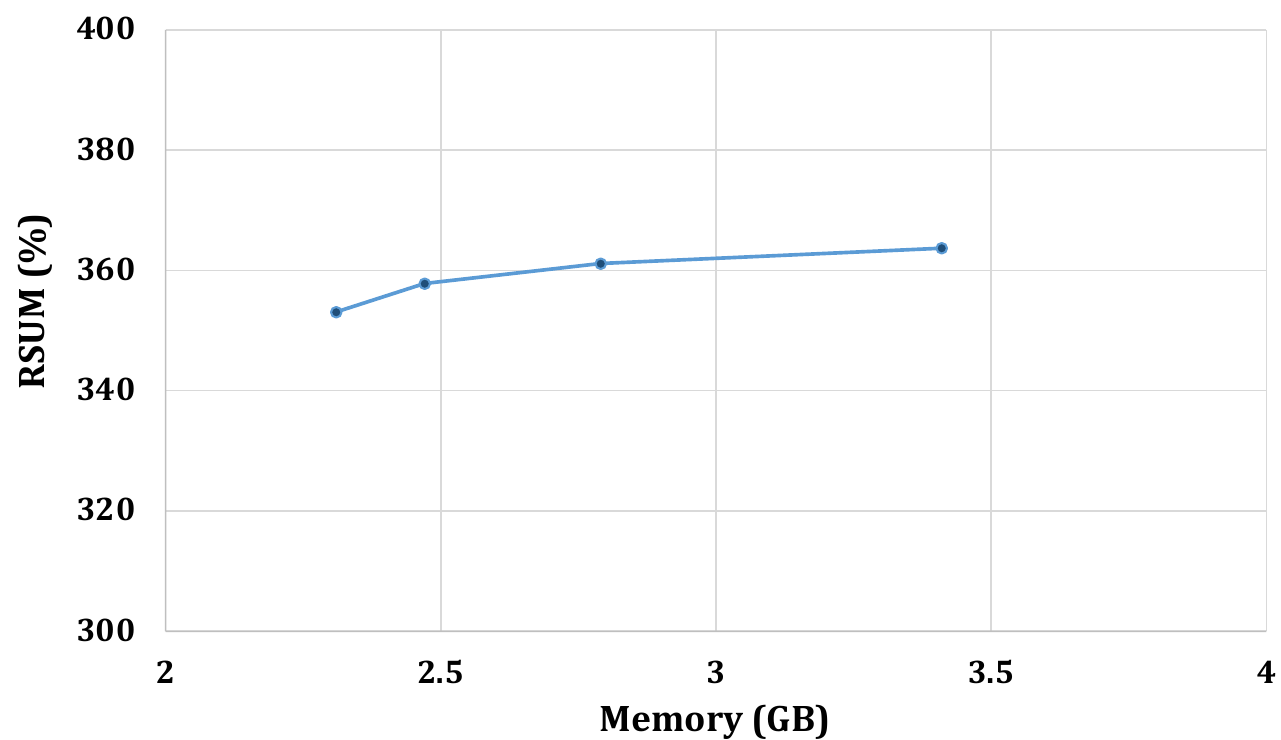}
        \caption{Accuracy-memory trade-off with varying reduction factor on the VTR task. Following UniPT on MSR-VTT, we search $r$ over $\{8,4,2,1\}$ using \emph{CLIP4Clip}~\cite{VLP:CLIP4Clip}.}
        \label{fig:reduction_factor_VTR}
    \end{minipage}
    \hfill%
    \begin{minipage}[t]{0.48\textwidth}
        \centering
        \includegraphics[width=\linewidth, height=0.65\linewidth,trim= 0 185 345 0,clip]{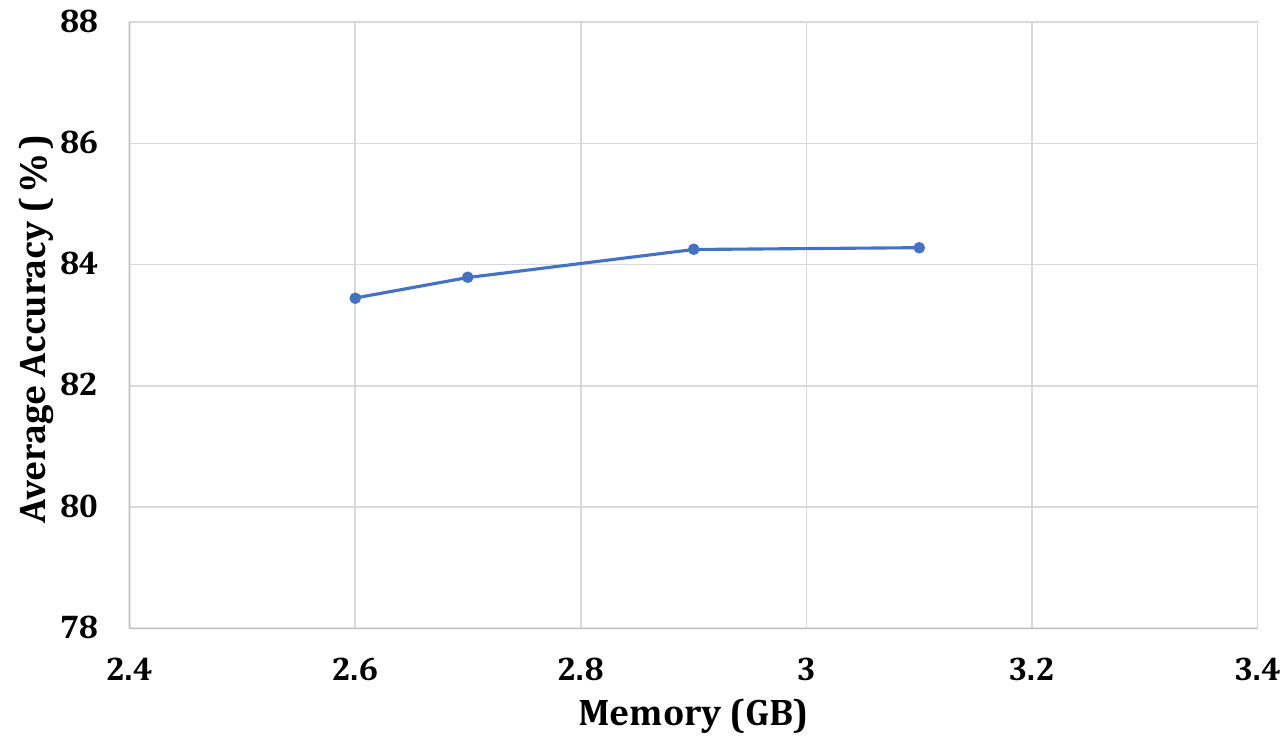}
        \caption{Accuracy-memory trade-off with varying reduction factor on NLP domain. Following LST on GLUE, we search $r$ over $\in \{32,16,8,4\}$ using \emph{T5-base}~\cite{TransF:T5}.}
        \label{fig:reduction_factor_GLUE}
        \end{minipage} 
\end{figure*}

\section{Hyper-parameter Reduction Factor}
\cref{fig:reduction_factor_VTR,fig:reduction_factor_GLUE} show the accuracy-memory trade-off via different reduction factor $r$. Following LST~\cite{TL:LST}, we search $r\in\{32,16,8,4\}$ on the GLUE benchmark, while for the VTR task, we follow UniPT~\cite{TL:UniPT} to adjust $r\in\{8,4,2,1\}$ on MSR-VTT. We discover that whether on uni-modal or cross-modal domains, our SHERL exhibits strong robustness and stability across varying reduction factor $r$. In line with early methods, we set $r$ as 8 for NLP tasks and 2 for VL tasks respectively, which strikes an optimal balance between memory usage and performance gains.    
\end{appendix}

\bibliographystyle{splncs04}
\bibliography{main}
\end{document}